\documentclass[paper=letter, twocolumn, 10pt]{scrartcl}
\pdfoutput=1
\usepackage{lmodern}				\usepackage{fullpage}
					\usepackage[latin1]{inputenc}		\usepackage[T1]{fontenc} 
\title{
	Multi-Information Source Optimization 
}
\date{}
\usepackage{authblk}
\author{Matthias Poloczek}
\author{Jialei Wang} 
\author{Peter I.\ Frazier}
\affil{School of Operations Research and Information Engineering\\Cornell University\\
	\texttt{\{poloczek,jw865,pf98\}@cornell.edu}}

\usepackage{natbib}
\usepackage[cmex10]{amsmath}
\usepackage{amssymb,amsthm,url}
\usepackage{bbm,bm}					
\usepackage{microtype}

\usepackage{xcolor}					\usepackage{hyperref}				\usepackage{graphicx}
\usepackage{complexity}
\usepackage{url}            \usepackage{booktabs}       
\usepackage{xspace}
\renewcommand{\E}{\ensuremath{\mathbb{E}}}		
\newcommand{\argmax}{\operatornamewithlimits{argmax}}

\newcommand{\I}{\ensuremath{\mathbbm{1}}}
\newcommand{\domain}{\ensuremath{{\cal D}}\xspace}
\newcommand{\discretedomain}{\ensuremath{{\cal A}}\xspace}
\newcommand{\EGO}{\texttt{EGO}\xspace}
\newcommand{\IS}{\ensuremath{{\cal IS}}\xspace}

\newcommand{\EI}{\texttt{EI}\xspace}
\newcommand{\MEI}{\texttt{misoEI}\xspace}
\newcommand{\MKG}{\texttt{misoKG}\xspace}
\newcommand{\ES}{\texttt{MTBO+}\xspace}

\allowdisplaybreaks

\begin{document}
	
	\maketitle

	\newcommand{\misotitle}{Multi-Information Source Optimization}

	\begin{abstract}
We consider Bayesian optimization of an expensive-to-evaluate black-box objective function, where we also have access to cheaper approximations of the objective.
In general, such approximations arise in applications such as reinforcement learning, engineering, and the natural sciences, and are subject to an inherent, unknown bias. This model discrepancy is caused by an inadequate internal model that deviates from reality and can vary over the domain, making the utilization of these approximations a non-trivial task.

We present a novel algorithm that provides a rigorous mathematical treatment of the uncertainties arising from model discrepancies and noisy observations. Its optimization decisions rely on a value of information analysis that extends the Knowledge Gradient factor to the setting of multiple information sources that vary in cost: each sampling decision maximizes the predicted benefit per unit cost.

We conduct an experimental evaluation that demonstrates that the method consistently outperforms other state-of-the-art techniques: it finds designs of considerably higher objective value and additionally inflicts less cost in the exploration process.
\end{abstract}

	\section{Introduction}
\label{section_intro}
We consider the problem of optimizing an expensive-to-evaluate black-box function, where we additionally have access to cheaper approximations of the objective.
For instance, this scenario arises when tuning hyper-parameters of machine learning algorithms, e.g., parameters of a neural network architecture or a regression method: one may assess the performance of a particular setting on a smaller related dataset or even a subset of the whole database (e.g., see~\cite{ssa13,kdosp16,kfbhh16}).
Or consider a ride-sharing company, e.g., Uber or Lyft, that matches clients looking for a ride to drivers. These real-time decisions then could be made by a system whose performance crucially depends on a variety of parameters. These parameters can be evaluated either in a simulation or by deploying them in a real-world experiment.
Similarly, the tasks of configuring the system that navigates a mobile robot in a factory, or steers a driverless vehicle on the road, can be approached by invoking a simulator, or by evaluating its performance on a closed test course or even a public road.
The approach of exploiting cheap approximations is fundamental to make such tasks tractable.

We formalize these problems as \emph{multi-information source optimization problems} (MISO), where the goal is to optimize a complex design under an expensive-to-evaluate black-box function. To reduce the overall cost we may utilize cheap approximate estimates of the objective that are provided by so-called ``information sources''. 
However, in the general settings we consider their output is not necessarily an unbiased estimator only subject to observational noise, but in addition may be inherently biased: 
the behavior of an autonomous car on a test course or the performance of a dispatch algorithm in a ride-sharing simulation will differ significantly from the real world.
Thus, inevitably we face a \emph{model discrepancy} which denotes an inherent inability to describe the reality accurately. 
We stress that this notion goes far beyond typical ``noise'' such as measurement errors or numerical inaccuracies. 
The latter grasps uncertainty that arises when sampling from an information source, and is the type of inaccuracy that most of the previous work on multifidelity optimization deals with, e.g., see~\cite{bv04,eldred2004second,rajnarayan2008multifidelity,mw12,law15} for problem formulations in the engineering sciences. 
In particular, such an understanding of noise assumes implicitly that the internal value (or state) of the information source itself is an accurate description of the truth.
Our more general notion of model discrepancy captures the uncertainty about the truth that originates from the inherent deviation from reality.

\paragraph*{Related Work.}
The task of optimizing a single, expensive-to-evaluate black-box function has received a lot of attention. A successful approach to this end is Bayesian optimization, a prominent representative being the efficient global optimization~(\EGO) method by \citet{JoScWe98}. 
\EGO assumes that there is only a single information source that returns the true value; the goal is to find a global maximum of the objective while minimizing the query cost.
It consists of the following two steps that have served as a blueprint for many subsequent Bayesian optimization techniques.
First a stochastic, in their case Gaussian, process is formulated and fitted to a set of initial data points. 
Then they search for a global optimum by iteratively sampling a point of highest predicted score according to some ``acquisition criterion'': Jones et al.\ employ \emph{expected improvement}~(\EI) that samples a point next whose  improvement over the best current design is maximum in expectation.
Subsequently, the expected improvement method was extended to deal with observational noise, e.g., see~\cite{HuAlNoZe06,sfp11,pgrc13}.
The extension of~\citeauthor{pgrc13} also dynamically controls the precision of a measurement, making it well suited for many multifidelity optimization problems. 

The strategy of reducing the overall cost of the optimization process by utilizing cheaper approximations of the objective function was successfully employed in a seminal paper by~\citet*{ssa13}, who showed that the task of tuning hyper-parameters for two classification problems can be sped up significantly by evaluating settings on a small sample instead of the whole database. 
To this end, they proposed a Gaussian process model to quantify the correlation between such an ``auxiliary task'' and the primary task, building on previous work on Gaussian process regression for multiple tasks in~\cite{bcw07,goovaerts97,tsj05}: their kernel is given by the tensor product~$K_t \otimes K_x$, where~$K_t$ (resp., $K_x$) denotes the covariance matrix of the tasks (resp., of the points).
Their acquisition criterion is a cost-sensitive formulation of Entropy Search~\citep{hs12,vvw09}: here one samples in each iteration a point that yields a maximum reduction in the uncertainty over the location of the optimum.

Besides, interesting variants of the MISO problem have been studied recently:
\citet{kdosp16} examined an alternative objective for multifidelity settings that asks to minimize the \emph{cumulative} regret over a series of function evaluations: besides classification problems they also presented an application where the likelihood of three cosmological constants is to be maximized based on Supernovae data.
\citet{kfbhh16} considered hyper-parameter optimization of machine learning algorithms over a large dataset~$D$. Supposing access to subsets of~$D$ of arbitrary sizes, they show how to exploit regularity of performance across dataset sizes to significantly speed up the optimization process for support vector machines and neural networks.

In engineering sciences the approach of building cheap-to-evaluate, approximate models for the real function, that offer different fidelity-cost trade-offs, is also known as ``surrogate modeling'' and has gained a lot of popularity (e.g., see the survey~\cite{qhsgvt05}).
\citet{kh00} introduced Gaussian process regression to multifidelity optimization to optimize a design given several computer codes that vary in accuracy and computational complexity.
Contrasting the related work discussed above, these articles consider model discrepancy, but impose several restrictions on its nature:
a common constraint in multifidelity optimization (e.g., see~\cite{kh00,bv04,eldred2004second,rajnarayan2008multifidelity,mw12,gc15}) is that information sources are required to form a \emph{hierarchy}, thereby limiting possible correlations among their outputs: in particular, once one has queried a high fidelity source for some point~$x$, then no further knowledge on~$g(x)$ can be gained by querying any other information source of lower fidelity (at any point).
A second frequent assumption is that information sources are \emph{unbiased}, admitting only (typically normally distributed) noise that further must be \emph{independent} across different sources.
\citet*{law15} addressed several of these shortcomings by a novel surrogate-based approach that requires the information sources to be neither hierarchical nor unbiased, and allows a more general notion of model discrepancy building on theoretical foundations by \citet{aw14}. 
Their model has a separate Gaussian process for each information source that in particular quantifies its uncertainty over the domain. Predictions are obtained by fusing that information via the method of \cite{win81}.
Then they apply the \EI acquisition function on these surrogates to first decide what design~$x^\ast$ should be evaluated next and afterwards select the respective information source to query~$x^\ast$; the latter decision is based on a heuristic that aims to balance information gain and query cost (see also Sect.~\ref{section_experiments}).

\paragraph*{Our Contributions.}
We present an approach to multi-information source optimization that allows exploiting cheap approximative estimates of the objective, while handling model discrepancies in a general and stringent way. Thus, we anticipate a wide applicability, including scenarios such as described above. 
We build on the common technique to capture the model discrepancy of each information source by a Gaussian process. However, we break through the separation and build a single statistical model that allows a uniform Bayesian treatment of the black-box objective function and the information sources.
This model improves on previous works in multifidelity optimization and in particular on~\cite{law15}, as it allows rigorously exploiting correlations across different information sources and hence enables us to reduce our uncertainty about \emph{all} information sources when receiving \emph{one} new data point, even if it originates from a source with lower fidelity. 
Thus, we obtain a more accurate estimate of the true objective function from each sample.
Note that this feature is also accomplished by the product kernel of~\cite{ssa13}. 
Their model differs in the assumption that the relationship of any pair of information sources is \emph{fixed} throughout the domain, whereas our covariance function is more flexible. Additionally, our model incorporates and quantifies the bias between the objective and any other source explicitly.

Another important contribution is an acquisition function that quantifies the uncertainty about the true objective and information sources, in particular due to model discrepancies and noise. 
To this end, we show how the Knowledge Gradient (KG) factor proposed by \citet*{FrPoDa_Correlated} can be computed efficiently in the presence of multiple information sources that vary in cost.
This \emph{cost-sensitive Knowledge Gradient} selects a pair~$(\ell,x)$ such that the \emph{simple regret} (i.e.\ the loss of the current best solution with respect to the unknown global optimum) relative to the specific query cost~$c_\ell(x)$ is minimized in expectation. 
Specifically, we show that the query cost can be incorporated as part of the objective in a natural way:
our policy picks a pair that offers an \emph{optimal trade-off} between the predicted simple regret and the corresponding cost.
Specifically, its choice is even provably one-step Bayes optimal in terms of this benefit per unit cost.
We regard it a conceptual advantage that the cost-sensitive KG factor can be computed analytically, whereas~\citet{ssa13} rely on Monte Carlo approximations (see also~\cite{hs12} for a discussion). \cite{law15} utilize a two-step heuristic as acquisition function.

We also demonstrate that our model is capable of handling information sources whose model discrepancies are interrelated in a more sophisticated way: in particular, we address the scenario of groups of information sources whose models deviate from the truth in a correlated fashion.
For instance, in silico simulations of physical or chemical processes might employ similar approximations whose deviations from the physical laws are thus correlated, e.g., finite element analyses with different mesh fineness by~\citet{hanm06} or calculations based on shared data sets.
Additionally, experiments conducted in the same location are exposed to the same environmental conditions or singular events, thus the outputs of these experiments might deviate from the truth by more than independent measurement errors. 
Another important factor is humans involved in the lab work, as typically workers have received the comparable training and may have made similar experiences during previous joint projects, which influences their actions and decisions.

	\section{The Model}
\label{section_model}
Each design (or point)~$x$ is specified by~$d$ parameters. Given some compact set~$\domain \subset \mathbb{R}^d$ of feasible designs, our goal is to find a best design under some continuous objective function~$g: \domain \to \mathbb{R}$, i.e.\ we want to find a design in~$\argmax_{x \in \domain} g(x)$.
Restrictions on~$\domain$ such as box constraints can be easily incorporated in our model.
We have access to~$M$ possibly biased and/or noisy information sources $\IS_1, \IS_2,\ldots,\IS_M$ that provide information about $g$.
Note that the~$\IS_\ell$ (with~$\ell \in [M]_0$) are also called "surrogates"; in the context of hyper-parameter tuning they are sometimes referred to as ``auxiliary tasks'' and~$g$ is the primary task.
We suppose that repeatedly observing $\IS_\ell(x)$ for some~$\ell$ and~$x$ provides independent and normally distributed observations with mean value~$f(\ell,x)$ and variance~$\lambda_\ell(x)$.
These sources are thought of as approximating $g$, with variable model discrepancy or bias $\delta_\ell(x) = g(x) - f(\ell,x)$.
We suppose that~$g$ can be observed directly without bias (but possibly with noise) and set~$\IS_0 = g$.
Each~$\IS_\ell$ is also associated with a query cost function~$c_\ell(x) : \; \domain \to \mathbb{R}^+$.
We assume that the cost function~$c_\ell(x)$ and the variance function~$\lambda_\ell(x)$ are both known and continuously differentiable.
In practice, these functions may either be provided by domain experts or may be estimated along with other model parameters from data (see Sect.~\ref{section_experiments}, the supplement, and \cite{rw06}).
Our motivation in having the cost and noise vary over the space of designs is that physical experiments may become difficult to conduct and/or expensive when environmental parameters are extreme. Moreover, simulations may be limited to certain specified parameter settings and their accuracy diminish quickly.

We now place a single Gaussian process prior on $f$ (i.e., on $g$ and the mean response from the~$M$ information sources).
Let $\mu : [M] \times \domain \mapsto \mathbb{R}$ be the mean function of this Gaussian process,
and $\Sigma : ([M] \times \domain)^2 \mapsto \mathbb{R}$ be the covariance kernel.
(Here, for any~$a \in \mathbb{Z}^+$ we use~$[a]$ as a shorthand for the set~$\{1,2,\ldots,a\}$, and further define~$[a]_0 = \{0,1,2,\ldots,a\}$.)
While our method can be used with an arbitrary mean function and positive semidefinite covariance kernel, we  provide two concrete parameterized classes of mean functions and covariance kernels that are useful for multi-information source optimization. 
Due to space constraints the second class is deferred to the supplement, where we also detail how to estimate the hyper-parameters of the mean function and the covariance kernel.

\paragraph*{Independent Model Discrepancy.}
\label{section_simple_model}
We first propose a parameterized class of mean functions $\mu$ and covariance functions $\Sigma$ derived by supposing that model discrepancies are chosen independently across information sources.  This first approach is appropriate when information sources are different in kind from each other and share no relationship except the fact that they are modeling a common objective.  We also propose a more general parameterized class that models correlation between model discrepancies, as is typical when information sources can be partitioned into groups, such that information sources within a group tend to agree more amongst themselves than they do with information sources in other groups. Due to space constraints, this class was deferred to the the online supplement.

We suppose here that $\delta_\ell$ for each $\ell>0$ was drawn from a separate independent Gaussian process, $\delta_\ell \sim GP(\mu_\ell, \Sigma_\ell)$.  We also suppose that $\delta_0$ is identically $0$, and that $f(0,\cdot) \sim GP(\mu_0, \Sigma_0)$, for some given $\mu_0$ and $\Sigma_0$.  We then define $f(\ell,x) = f(0,x) + \delta_\ell(x)$ for each $\ell$.
Typically, one would not have a strong prior belief as to the direction of the bias inherent in an information source, and so we set $\mu_\ell(x) = 0$.  (If one does have a strong prior opinion that an information source is biased in one direction, then one may instead set $\mu_\ell$ to a constant estimated using maximum a posteriori estimation.)
With this modeling choice, we see that the mean of $f \sim GP(\mu,\Sigma)$ with mean function~$\mu$ and covariance kernel~$\Sigma$ is given by
\begin{equation*}
\mu(\ell,x) = \E\left[f(\ell,x)\right] = \E\left[f(0,x)\right] + \E\left[\delta_\ell(x)\right] \\
= \mu_0(x)
\end{equation*}
for each~$\ell \in [M]_0$, since~$\E\left[\delta_\ell(x)\right] = 0$ holds.
Additionally,
for~$\ell,m \in [M]_0$ and~$x,x' \in \domain$,
\begin{align*}
& \Sigma\left((\ell,x),(m,x')\right)\\
= \; & \mathrm{Cov}(f(0,x) + \delta_\ell(x), f(0,x') + \delta_m(x'))\\
= \; & \Sigma_0(x,x') + \I_{\ell,m} \cdot \Sigma_\ell(x,x'),
\end{align*}
where~$\I_{\ell,m}$ denotes Kronecker's delta, and where we have used independence of $\delta_\ell$, $\delta_m$, and $f(0,\cdot)$.

	\section{The Value of Information Analysis}
\label{section_voi}
Our optimization algorithm proceeds in rounds, where in each round it selects a design~$x \in \domain$ and an information source~$\IS_\ell$ with~$\ell \in [M]_0$. The goal is to find an~$x$ that maximizes~$g(x)$ over~$\domain$. 
Let us assume for the moment that query costs are uniform over the whole domain and all information sources; we will show how to remove this assumption later.
Further, assume that we have already sampled~$n$ points~$X$ and made the observations~$Y$.
Finally, denote by~$\mathbb{E}_n\left[f(\ell,x)\right]$ the expected value according to the posterior distribution given~$X$ and~$Y$ and shorthand~$\mu^{(n)}\left(\ell,x\right) := \mathbb{E}_n\left[f(\ell,x)\right]$. 
Since that distribution is normal, the best \emph{expected} objective value of any design, as estimated by our statistical model, is~$\max_{x' \in \domain} \mu^{(n)}\left(0,x'\right)$.
If we were to pick an~$x \in \domain$ now irrevocably, then we would select an~$x$ of maximum expectation.
This motivates choosing the next design~$x^{(n+1)}$ and information source~$\ell^{(n+1)}$ that we will sample such that we maximize
$\mathbb{E}_n\left[\max_{x' \in \domain} \mu^{(n+1)}(0,x') \;\middle|\; \ell^{(n+1)} = \ell,x^{(n+1)} = x\right]$, or equivalently maximize the expected \emph{gain} over the current optimum by
$\mathbb{E}_n\left[\max_{x' \in \domain} \mu^{(n+1)}(0,x') \;\middle|\; \ell^{(n+1)} = \ell,x^{(n+1)} = x\right] - \max_{x' \in \domain} \mu^{(n)}(0,x')$.  
Note that the equivalence of the maximizers for both expressions follows immediately from the observation that~$\mu^{(n)}(0,x')$ is a constant for all~$x' \in \domain$ given~$X$ and~$Y$.

Next we show how the assumption made at the beginning of this section, that query costs are uniform across the domain and for all information sources, can be removed.
To this end, we associate a query cost function~$c_\ell(x):  \domain \to \mathbb{R}^+$ with each information source~$\IS_\ell$ for~$\ell \in [M]_0$.
Then our goal becomes to find a sample~$(\ell^{(n+1)},x^{(n+1)})$ whose value of information divided by its respective query cost is maximum. 
The gist is that conditioned on any~$(\ell^{(n+1)},x^{(n+1)})$, the expected gain of all~$x' {\in} \domain$ is scaled by~$c_{\ell^{(n+1)}}(x^{(n+1)})^{-1}$.
Then the \emph{cost-sensitive Knowledge Gradient} policy picks a sample~$(\ell,x)$ that maximizes the expectation
\begin{multline}
\label{Eq_KG_cost}
\E_n\left[\frac{\max_{x' \in \domain}{\mu^{(n+1)}(0,x')} - \max_{x' \in \domain}{\mu^{(n)}(0,x')}}{c_{\ell}(x)} \;\middle|\; \right. \\
\left. \ell^{(n+1)} = \ell,x^{(n+1)} = x\right],
\end{multline}
denoted by~$\text{CKG}(\ell,x)$.
Our task is to compute~$(\ell^{(n+1)},x^{(n+1)}) \in \argmax_{\ell \in [M]_0, x \in \domain} \text{CKG}(\ell,x)$,
which is a nested optimization problem.

To make this task feasible in practice, we discretize the domain of the inner maximization problem stated in Eq.~(\ref{Eq_KG_cost}): for simplicity, we choose the discrete set~$\discretedomain \subset \domain$ via a Latin Hypercube design.
Alternatively, one could scatter the points to emphasize promising areas of~$\domain$ and resample regularly.
Now we have reduced the inner optimization problem for each information source to the setting of \citet{FrPoDa_Correlated} who showed how to compute the value of information over a discrete set if there is only one information source and query costs are uniform.

We provide an outline and refer to their article for details.
For their setting let~$\bar{\mu}^{n}$ be the vector of posterior means for~$\discretedomain$ and define for each~$x \in \discretedomain$ $\bar{\sigma}^n(x) = \Sigma^n e_x / (\lambda(x) + \Sigma^n_{xx})$, where~$\Sigma^n$ is the posterior covariance matrix and~$e_x \in \{0,1\}^{|\discretedomain|}$ with a one for~$x$ and zeros elsewhere.
Given these vectors, observe that
\begin{align*}
& \E_n\left[\max_{x' \in \discretedomain}{\mu^{(n+1)}(0,x')} - \max_{x' \in \discretedomain}{\mu^{(n)}(0,x')} \;\middle|\; x^{(n+1)} = x\right]\\
& = h\left(\bar{\mu}^{n},\bar{\sigma}^n(x)\right),
\end{align*}
where~$h(a,b) = \mathbb{E}_n \left[\max_i a_i + b_i Z\right] - \max_i a_i$ for vectors~$a,b$, and~$Z$ is a  one-dimensional standard normal random variable.
Frazier et al.\ show how to compute~$h$.

Thus, following our initial considerations, we approximate the cost-sensitive Knowledge Gradient by maximizing~$\frac{h\left(\bar{\mu}^{n},\bar{\sigma}^n(\ell,x)\right)}{c_{\ell}(x)}$ over~$[M]_0 \times \domain$, i.e.\ the outer optimization problem is still formulated over~$\domain$.
Note that we can compute the gradient of~$\frac{h\left(\bar{\mu}^{n},\bar{\sigma}^n(\ell,x)\right)}{c_{\ell}(x)}$ with respect to~$x$ assuming that~$c_{\ell}$ is differentiable (e.g., when given by a suitable Gaussian process). 
Thus, we may apply a multi-start gradient-based optimizer to compute $(\ell^{(n+1)},x^{(n+1)})$.

We summarize our algorithm~\MKG:
\begin{enumerate}
\item Using samples from all information sources, estimate hyper-parameters of the Gaussian process prior as described in the online supplement.

Then calculate the posterior~$f$ based on the prior and samples.
	
\item Until the budget for samples is exhausted do:

Determine the information source~$\ell {\in} [M]_0$ and the design~$x {\in} \domain$ that maximize the cost-sensitive Knowledge Gradient proposed in Eq.~(\ref{Eq_KG_cost}) and observe~$\IS_\ell(x)$.

Update the posterior distribution with the new observation.

\item Return the point with the largest estimated value according to the current posterior~$f(0,\cdot)$.
\end{enumerate}

	\section{Numerical Experiments}
\label{section_experiments}
We demonstrate the performance of the proposed multi-information source optimization algorithm, \MKG, by comparing it with the state-of-the-art Bayesian optimization algorithms for MISO problems.
The statistical model and the value of information analysis were implemented in \texttt{Python~2.7} and~\texttt{C{++}}  using the functionality provided by the \texttt{Metrics Optimization Engine}~\cite{moe-github2015}.
\paragraph*{Benchmark Algorithms.} 

The first benchmark method, \ES, is an improved version of Multi-Task Bayesian Optimization proposed by \citet{ssa13}. It uses a cost-sensitive version of Entropy Search as acquisition function that picks samples to maximize the information gain over the location of the optimum of the objective function, normalized by the respective query cost (see their paper for details). 
\texttt{MTBO} combines acquisition function with a ``multi-task'' Gaussian process model that captures the relationships between information sources (the ``tasks'') and the objective function.
Following a recommendation of Snoek~\citeyear{snoek16}, our implementation \ES uses an improved formulation of the acquisition function given by~\citet{hhg14,spearmint_github}, 
but otherwise is identical to~\texttt{MTBO}; in particular, it uses the statistical model of~\citet{ssa13}. 

The other algorithm,~\MEI of~\cite{law15}, was developed to solve MISO problems that involve model discrepancy and therefore is a good competing method to compare with. 
It maintains a separate Gaussian process for each information source: to combine this knowledge, the corresponding posterior distributions are fused for each design via Winkler's method (\citeyear{win81}) into a single intermediate surrogate, which is a normally distributed random variable.
Then Lam et al.\ adapt the Expected Improvement (\EI) acquisition function to select the design which is to be sampled next:
for the sake of simplicity, assume that observations are noiseless and that~$y^\ast$ is the objective value of a best sampled design. If~$Y_x$ denotes a Gaussian random variable with the posterior distribution of the objective value for design~$x$, then~$\E[\max\{Y_x - y^\ast, 0\}]$ is the expected improvement for~$x$, and the~\EI acquisition function selects an~$x$ that maximizes this expectation.
Based on this decision, the information source to invoke is chosen by a heuristic that aims at maximizing the~\EI per unit cost.

\paragraph*{The Experimental Setups.} We conducted numerical experiments on the following test problems: the first is the 2-dimensional Rosenbrock function which is a standard benchmark in the literature, tweaked into the MISO setting by~\citet{law15}. 
The second is a MISO benchmark proposed by~\citet{ssa13}: the goal is to optimize the four hyper-parameters of a machine learning algorithm, using a small, related set of smaller images as cheap information source.
The third is an assemble-to-order problem introduced by \citet{hong2006discrete}: here the objective is to optimize an~$8$-dimensional target stock vector in order to maximize the expected daily profit of a company, for which an estimate is provided as an output by their simulator.

In MISO settings the amount of initial data that one can use to inform the methods about each information source is typically dictated by the application, in particular by resource constraints and the availability of the respective source. 
In our experiments all methods were given \emph{identical initial datasets} for each information source in every replication; these sets were drawn randomly via Latin Hypercube designs.
For the sake of simplicity, we provided the same number of points for each~\IS, deliberately set in advance to 2.5 points per dimension of the design space~$\domain$. 
Regarding the kernel and mean function, \ES uses the settings provided in~\citep{spearmint_github}. The other algorithms used the squared exponential kernel and a constant mean function set to the average of a random sample.

We report the ``gain'' over the best initial solution, that is the true objective value of the respective design that a method would return at each iteration minus the best value in the initial data set. 
If the true objective value is not known for a given design, we report the value obtained from the information source of highest fidelity.
This gain is plotted as a function of the \emph{total cost}, that is the cumulative cost for invoking the information sources plus the fixed cost for the initial data; this metric naturally generalizes the number of function evaluations prevalent in Bayesian optimization.
Note that the computational overhead of choosing the next information source and sample is omitted, as it is negligible compared to invoking an information source in real-world applications.
Error bars are shown at the mean plus and minus \emph{two standard errors} averaged over at least~100 runs of each algorithm.
Note that even for deterministic sources a tiny observational noise of~$10^{-6}$ is supposed to avoid numerical issues during matrix inversion.

\subsection{The Rosenbrock Benchmarks}
We consider the design space~$\domain = [-2,2]^2$, and $M=2$ information sources.
$\IS_0$ is the Rosenbrock function~$f(\cdot)$ plus optional Gaussian noise, and
$\IS_1$ equals~$f(\cdot)$ with an additional oscillatory component:
\begin{equation} \label{eq:rosenbrock}
\begin{split}
&f(\bm{x}) = (1-x_1)^2 + 100 \cdot (x_2 - x_1^2)^2 \\
&\IS_0(\bm{x}) = f(\bm{x}) + u \cdot \varepsilon \\
&\IS_1(\bm{x}) = f(\bm{x}) + v \cdot \sin(10 \cdot x_1 + 5 \cdot x_2)
\end{split}
\end{equation}
where $\bm{x} = (x_1,x_2)^T \in \domain$, and $\varepsilon$ is i.i.d.\ noise drawn from the standard normal distribution. 
$u$ and $v$ are configuration constants.
We suppose that~$\IS_1$ is not subject to observational noise, hence the uncertainty only originates from the model discrepancy.
We experimented on two different configurations to gain a better insight into characteristics of the algorithms.
Since Lam et al.\ reported a good performance of their method on~\eqref{eq:rosenbrock}, we replicated their experiment using the same parameters to compare the performance of the four methods: 
that is, we set~$u=0$, $v = 0.1$. 
Replicating the setting in~\citep[p.~15]{law15}, we also suppose a tiny uncertainty for~$\IS_0$, although it actually outputs the truth, and set~$\lambda_0(x) = 10^{-3}$ and~$\lambda_1(x) = 10^{-6}$ for all~$x$.
Furthermore, we assume a cost of~$1000$ for each query to~$\IS_0$ and of~$1$ for~$\IS_1$.

\begin{figure} 
\includegraphics[width=\linewidth]{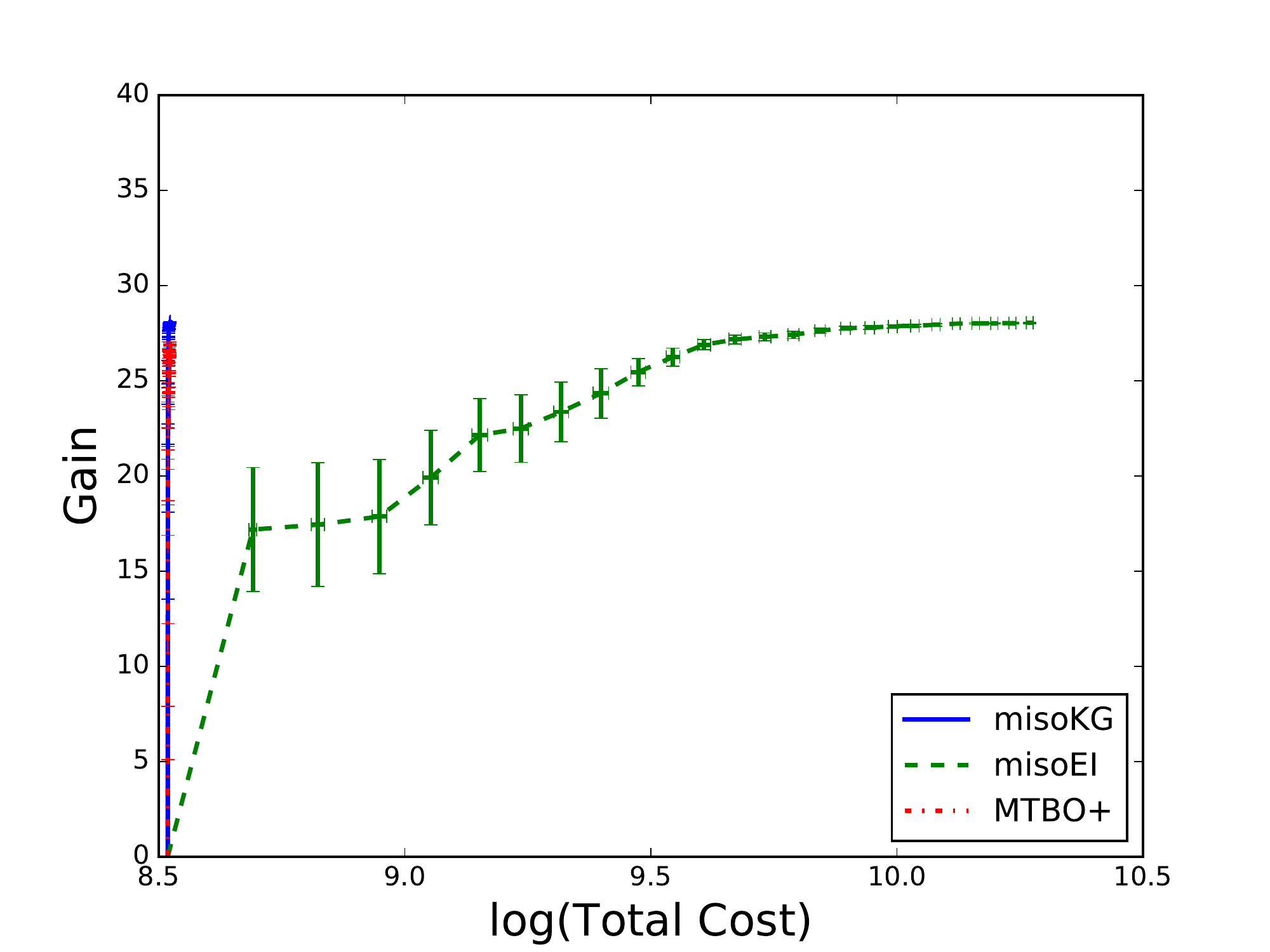} 
\includegraphics[width=\linewidth]{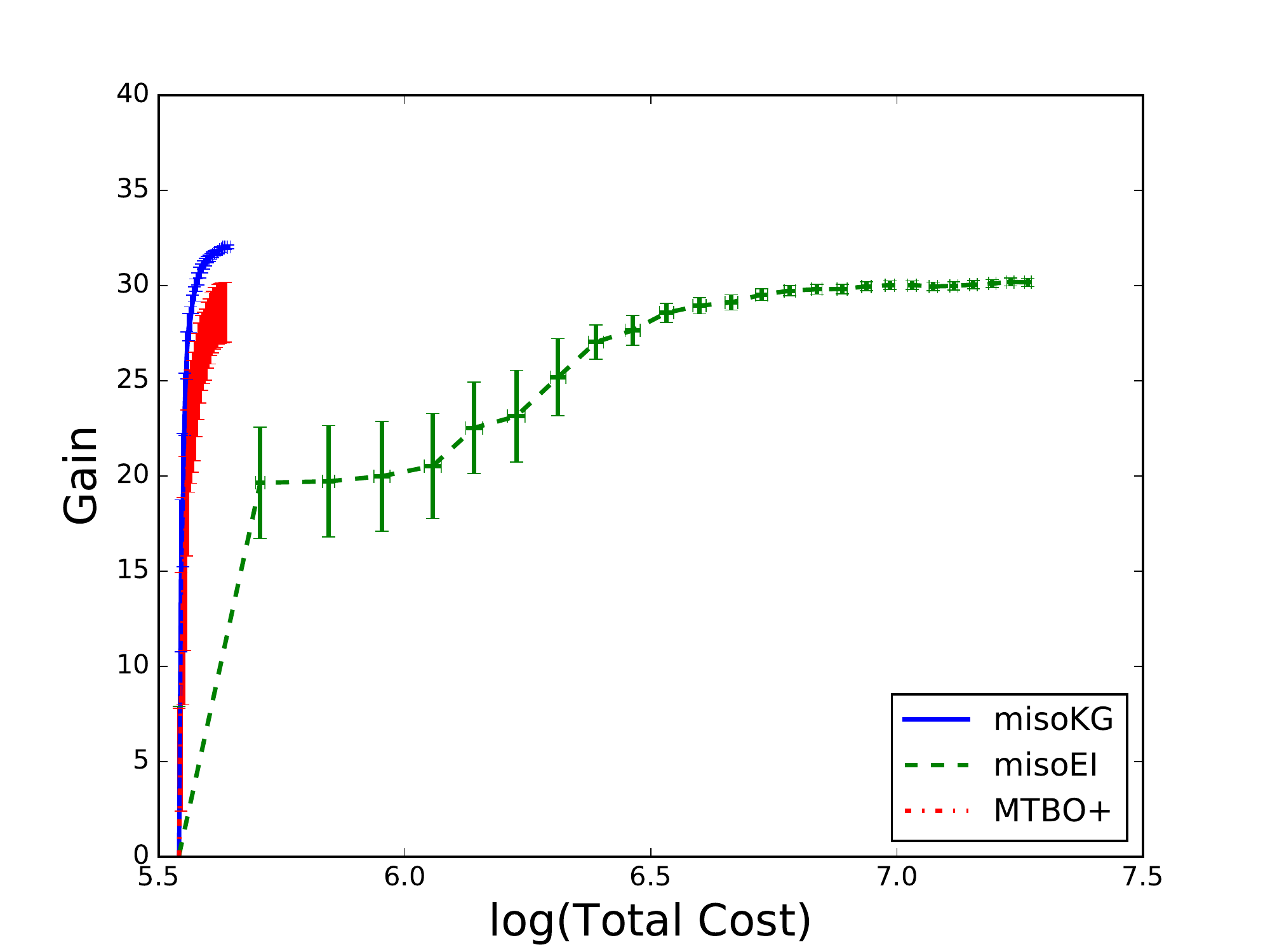} 
\caption{(t)  The Rosenbrock benchmark with the parameter setting of~\cite{law15}. (b) The Rosenbrock benchmark with the alternative setup.  \MKG offers an excellent gain-to-cost ratio and outperforms its competitors substantially.}
\label{fig_rb_1}
\label{fig_rb_2}
\end{figure}

Since all methods converged to good solutions within few queries, we investigate the ratio of gain to cost: Fig.~\ref{fig_rb_1}~(t) displays the gain of each method over the best initial solution as a function of the total cost, that is the cost of the initial data and the query cost accumulated by the acquisition functions.
We see that the new method~\MKG offers a better gain per unit cost, typically finding an almost optimal solution within~$5-10$ samples.
We note that~\MKG relies only on cheap samples, therefore managing the uncertainties successfully.
This is also true for~\ES that obtains a slightly worse solution. 
\MEI on the other hand queries the expensive truth to find the global optimum, thereby accumulating considerably higher cost.

For the second setup, we make the following changes: we set~$u=1$ and~$v=2$, and suppose for~$\IS_0$ uniform observational noise of~$\lambda_0(x) = 1$ and uniform query cost~$c_0(x) = 50$.
Note that now the difference of the costs of both sources is much smaller, while their uncertainties are considerably bigger.
The results are displayed in Fig.~\ref{fig_rb_2}~(b): 
as for the first configuration, \MKG outperforms the other methods, making efficient use of the cheap biased information source. 
In fact, the relative difference in performance is even larger, which might suggest that \MKG handles the observational noise slightly better than its competitors for this benchmark.

\subsection{The Image Classification Benchmark}
\label{section_lrMU}
This classification problem was introduced by \citet{ssa13} to demonstrate the performance of their~\texttt{MTBO} algorithm. 
The goal is to optimize hyper-parameters of the Logistic Regression algorithm~\cite{logregTheano} in order to minimize the classification error on the MNIST dataset of~\citet{mnist}. 
The weights are trained using a stochastic gradient method with mini-batches. We have four hyper-parameters: the learning rate, the L2-regularization parameter, the batch size, and the number of epochs.
The MNIST dataset contains~70,000 grayscale images of handwritten digits, where each image consists of~784 pixels.
In the experimental setup information source~$\IS_0$ corresponds to invoking the machine learning algorithm on this dataset.

Following Swersky et al., the USPS dataset serves as cheap information source~$\IS_1$: this set comprises only about~9000 images of handwritten digits that are also smaller, only~256 pixels each~\cite{uspsdata}. 
Again we suppose a tiny observational noise of~$10^{-6}$ and set the invocation costs of the sources to~4.5 for~$\IS_1$ and~43.69 for~$\IS_0$.
A closer examination shows that~$\IS_1$ is subject to considerable bias with respect to~$\IS_0$, making it a challenge for MISO algorithms.

\begin{figure} \centering

\includegraphics[width=\linewidth]{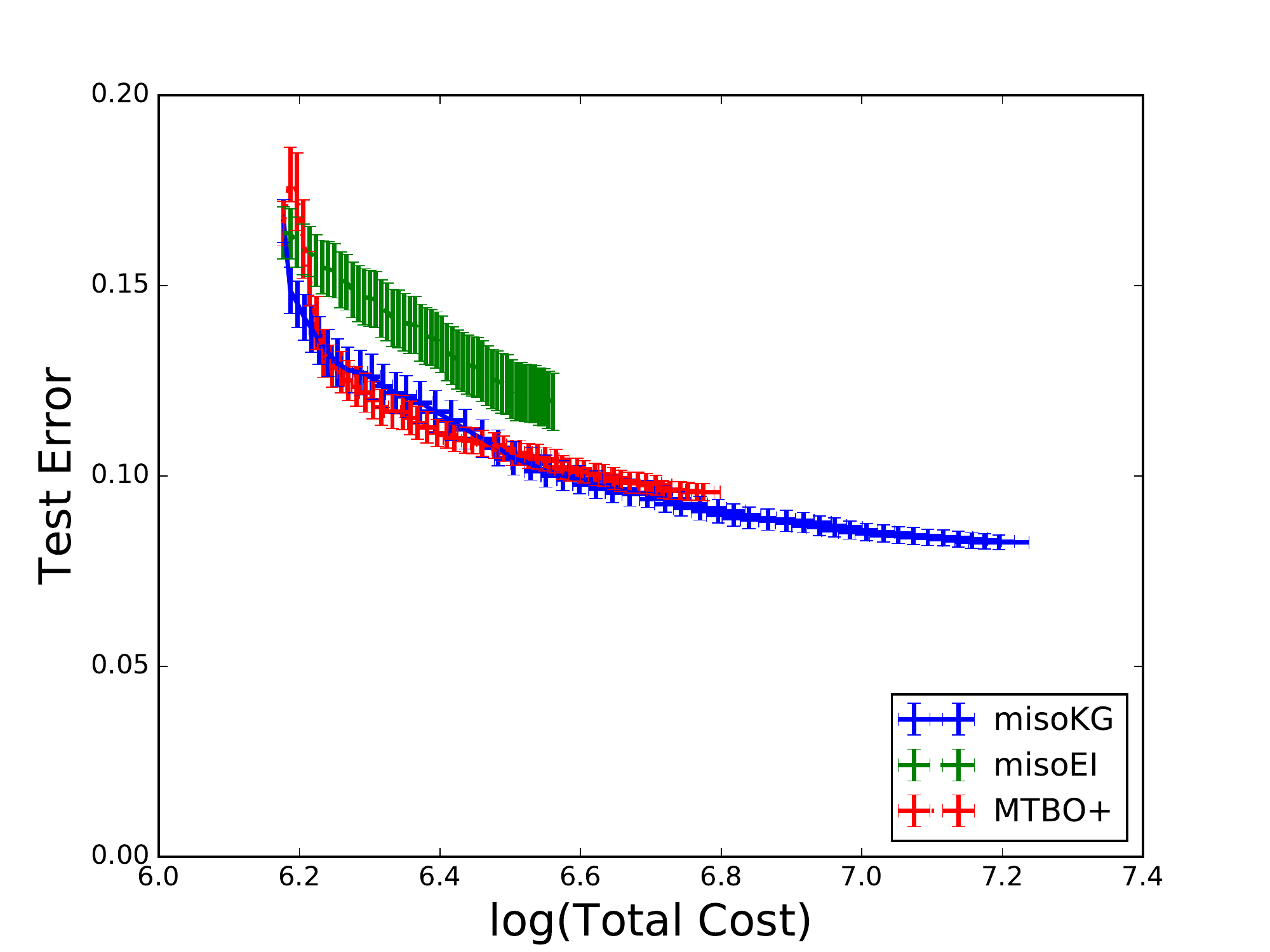}

\includegraphics[width=\linewidth]{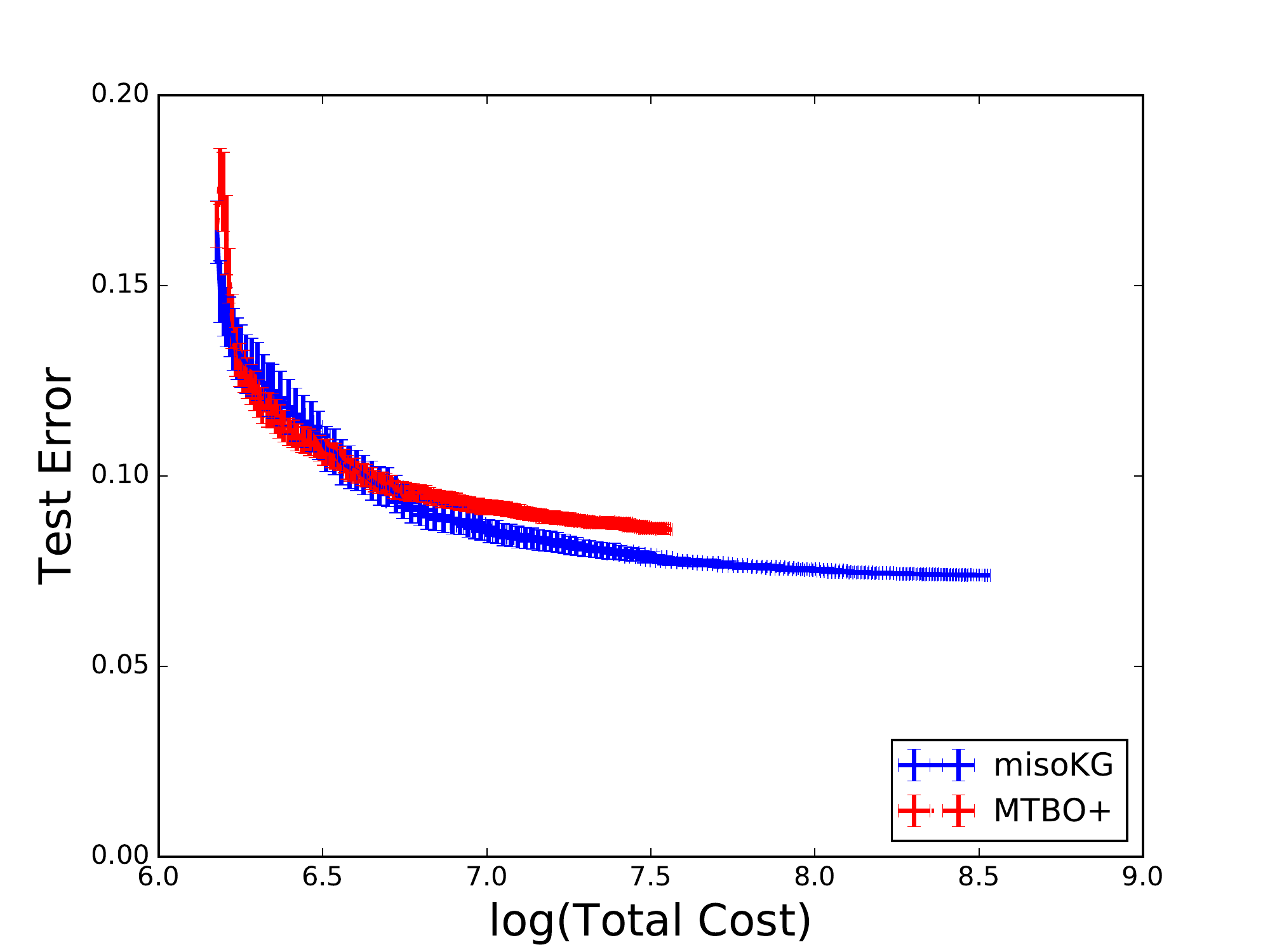}

\caption{
The performance on the image classification benchmark with significant model discrepancy.
(t) The first~50 steps of each algorithm: \MKG and \ES perform better than~\MEI.
(b) The first~150 steps of \MKG and \ES. While the initial performance of \MKG and \ES is comparable, \MKG achieves better testerrors after about 80 steps and converges to the global optimum.}
\label{fig_lrMU}
\end{figure}

Initially, \MKG and \ES are on par and both outperform~\MEI (cp.\ Fig.\ref{fig_lrMU} (t)). 
In order to study the convergence behavior, we evaluated~\MKG and~\ES for 150 steps, with a lower number of replications but using the same initial data for each replication. 
We observe that~\MKG usually achieves an optimal testerror of about~$7.1\%$ on the MNIST testset after about 80 queries to information sources (see Fig.\ref{fig_lrMU} (b)). 
In its late iterations \ES achieves a worse performance than \MKG has at the same costs.
Note that the experimental results of~\citet{ssa13} show that \ES will also converge to the optimum eventually.

\subsection{The Assemble-To-Order Benchmark}
In the assemble-to-order (ATO) benchmark, a reinforcement learning problem from a business application, we are managing the inventory of a company that manufactures~$m$ products. Each of these products is made from a selection from~$n$ items, where we distinguish for each product between key items and non-key items: if the company runs out of key items, then it cannot sell the respective products until it has restocked its inventory; non-key items are optional and used if available.
When the company sends a replenishment order, the required item is delivered after a random period whose distribution is known. 
Since items in the inventory inflict holding cost, our goal is to find an optimal target inventory
level vector~$b$ that determines the amount of each item we want to stock, such that we maximize the expected profit per day (cp.~\citet{hong2006discrete} for details).

Hong and Nelson proposed a specific scenario with~$m{=}5$ different products that depend on a subset of~$n{=}8$ items, thus our task is to optimize the~$8$-dimensional target vector~$b \in [0,20]^8$.
For each such vector their simulator provides an estimate of the expected daily profit by running the simulation for a variable number of replications (see ``runs'' in Table~\ref{table_ATO_parameters}). Increasing this number yields a more accurate estimate but also has higher computational cost.
The observational noise and query cost, i.e.\ the computation time of the simulation, are estimated from samples for each information source, assuming that both functions are constant over the domain for the sake of simplicity.

There are two simulators for this assemble-to-order setting that differ subtly in the model of the inventory system. However, the effect in estimated objective value is significant: on average the outputs of both simulators at the same target vector differ by about~$5\%$ of the score of the global optimum, which is about~$120$,  
whereas the largest observed bias out of~$1000$ random samples was~$31.8$.
Moreover, the sample variance of the difference between the outputs of both simulators is about~$200$. Thus, we are witnessing a significant model discrepancy.
We set up three information sources (cp.\ Table~\ref{table_ATO_parameters}): $\IS_0$ and~$\IS_2$ use the simulator of~\citet{simopt_ato}, whereas the cheapest source~$\IS_1$ invokes the implementation of~\citeauthor{hong2006discrete}.
We assume that~$\IS_0$ models the truth.
\begin{figure} \centering

\includegraphics[width=\linewidth]{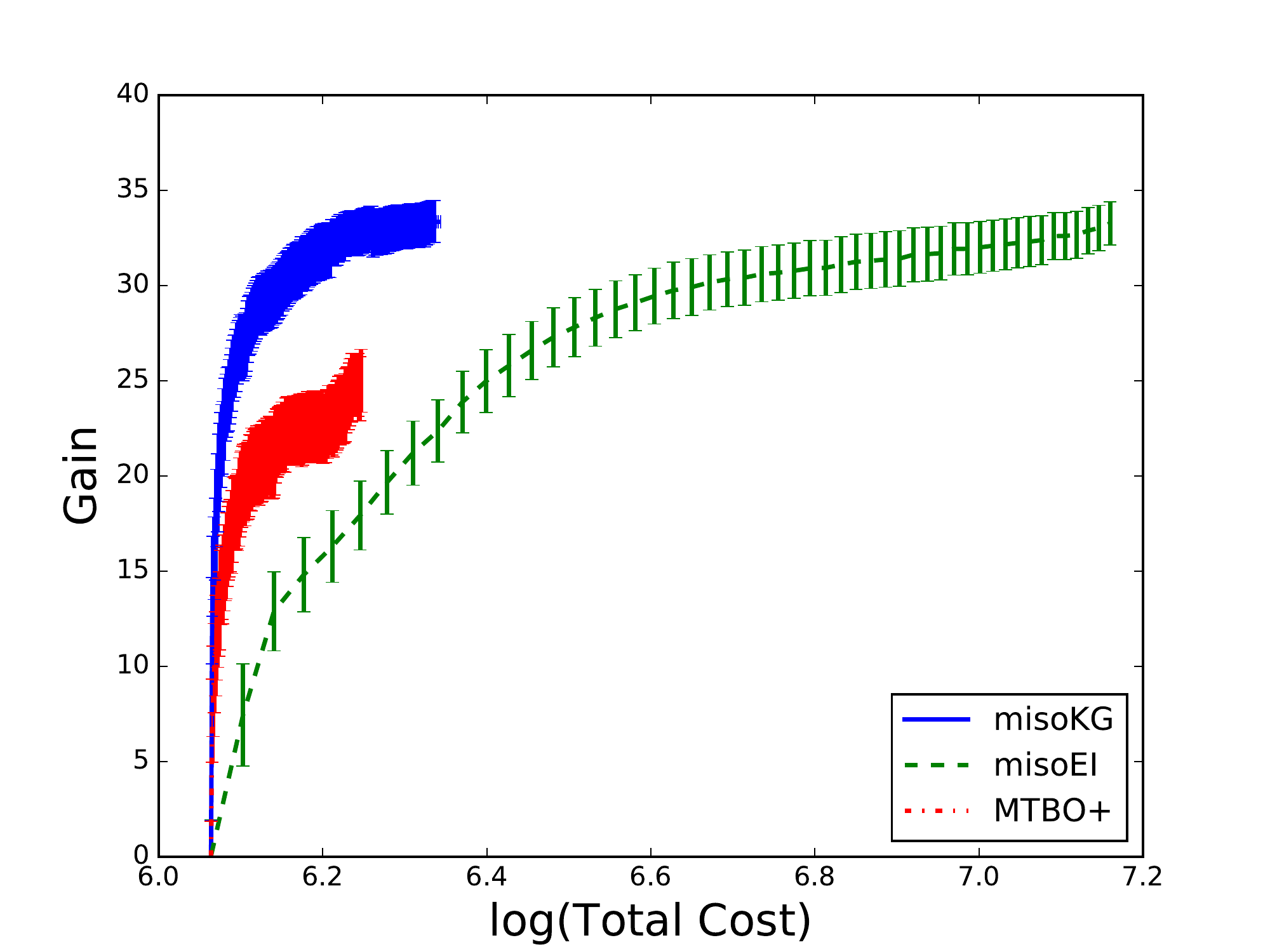}

\caption{The performance on the assemble-to-order benchmark with significant model discrepancy. \MKG has the best gain per cost ratio among the algorithms.}
\label{fig_atoext}
\end{figure}
\begin{table}[h]
\centering
\caption{The Parameters for the ATO problem}
\label{table_ATO_parameters}\begin{tabular}{rrrr}
& \# Runs & Noise Variance & Cost \\
\hline \\
$\IS_0$ & $500$ & $0.056$ & $17.1$\\
$\IS_1$ & $10$ & $2.944$ & $0.5$\\
$\IS_2$ & $100$ & $0.332$ & $3.9$\\
\end{tabular}\end{table}
Fig.~\ref{fig_atoext} displays the performance over the first~150 steps for~\MKG and~\ES and the first 50 steps of~\MEI, all averaged over~100 runs. 
\MKG has the best start and dominates in particular~\ES clearly.
\MKG averages at a gain of~$26.1$, but inflicts only an average query cost of~$54.6$ to the information sources, excluding the fixed cost for the initial datasets that are identical for all algorithms for the moment. This is only~$6.3\%$ of the query cost that~\MEI requires to achieve a comparable score.
Interestingly, \MKG and~\ES utilize in their optimization of the target inventory level vector mostly the cheap,  biased source, and therefore are able to obtain significantly better gain per cost ratios than~\MEI. 

Looking closer, we see that typically \MKG's first call to~$\IS_2$ occurs after about~$60-80$ steps. In total, \MKG queries~$\IS_2$ about ten times within the first~$150$ steps; in some replications \MKG makes one late call to~$\IS_0$ when it has already converged.
Our interpretation is that~\MKG exploits the cheap, biased~$\IS_1$ to zoom in on the global optimum and switches to the unbiased but noisy~$\IS_2$ to identify the optimal solution exactly.
This is the expected (and desired) behavior for~\MKG when the uncertainty of~$f(0,x^\ast)$ for some~$x^\ast$ is not expected to be reduced sufficiently by queries to~$\IS_1$.

\ES trades off the gain versus cost differently: it queries~$\IS_0$ once or twice after~$100$ steps and directs all other queries to~$\IS_1$, which might explain the observed lower performance.
\MEI that employs a two-step heuristic for trading off predicted gain and query cost chose almost always the most expensive simulation to evaluate the selected design.

\subsection*{Acknowledgments}
The authors were partially supported by NSF CAREER CMMI-1254298, NSF CMMI-1536895, NSF IIS-1247696, AFOSR FA9550-12-1-0200, AFOSR FA9550-15-1-0038, and AFOSR FA9550-16-1-0046.

	
		\bibliography{multifidelity}			

\begin{thebibliography}{34}
\providecommand{\natexlab}[1]{#1}
\providecommand{\url}[1]{\texttt{#1}}
\expandafter\ifx\csname urlstyle\endcsname\relax
  \providecommand{\doi}[1]{doi: #1}\else
  \providecommand{\doi}{doi: \begingroup \urlstyle{rm}\Url}\fi

\bibitem[Allaire and Willcox(2014)]{aw14}
D.~Allaire and K.~Willcox.
\newblock A mathematical and computational framework for multifidelity design
  and analysis with computer models.
\newblock \emph{International Journal for Uncertainty Quantification},
  4\penalty0 (1), 2014.

\bibitem[Balabanov and Venter(2004)]{bv04}
V.~Balabanov and G.~Venter.
\newblock Multi-fidelity optimization with high-fidelity analysis and
  low-fidelity gradients.
\newblock In \emph{10th AIAA/ISSMO Multidisciplinary Analysis and Optimization
  Conference}, 2004.

\bibitem[Bonilla et~al.(2007)Bonilla, Chai, and Williams]{bcw07}
E.~V. Bonilla, K.~M. Chai, and C.~Williams.
\newblock Multi-task gaussian process prediction.
\newblock In \emph{Advances in Neural Information Processing Systems}, pages
  153--160, 2007.

\bibitem[Eldred et~al.(2004)Eldred, Giunta, and Collis]{eldred2004second}
M.~S. Eldred, A.~A. Giunta, and S.~S. Collis.
\newblock Second-order corrections for surrogate-based optimization with model
  hierarchies.
\newblock In \emph{Proceedings of the 10th AIAA/ISSMO Multidisciplinary
  Analysis and Optimization Conference}, pages 2013--2014, 2004.

\bibitem[Frazier et~al.(2009)Frazier, Powell, and Dayanik]{FrPoDa_Correlated}
P.~I. Frazier, W.~B. Powell, and S.~Dayanik.
\newblock {The Knowledge Gradient Policy for Correlated Normal Beliefs}.
\newblock \emph{INFORMS Journal on Computing}, 21\penalty0 (4):\penalty0
  599--613, 2009.

\bibitem[Goovaerts(1997)]{goovaerts97}
P.~Goovaerts.
\newblock \emph{Geostatistics for Natural Resources Evaluation}.
\newblock Oxford University, 1997.

\bibitem[Gratiet and Cannamela(2015)]{gc15}
L.~L. Gratiet and C.~Cannamela.
\newblock Cokriging-based sequential design strategies using fast
  cross-validation techniques for multi-fidelity computer codes.
\newblock \emph{Technometrics}, 57\penalty0 (3):\penalty0 418--427, 2015.

\bibitem[Hennig and Schuler(2012)]{hs12}
P.~Hennig and C.~J. Schuler.
\newblock Entropy search for information-efficient global optimization.
\newblock \emph{The Journal of Machine Learning Research}, 13\penalty0
  (1):\penalty0 1809--1837, 2012.

\bibitem[Hern{\'a}ndez-Lobato et~al.(2014)Hern{\'a}ndez-Lobato, Hoffman, and
  Ghahramani]{hhg14}
J.~M. Hern{\'a}ndez-Lobato, M.~W. Hoffman, and Z.~Ghahramani.
\newblock Predictive entropy search for efficient global optimization of
  black-box functions.
\newblock In \emph{Advances in Neural Information Processing Systems}, pages
  918--926, 2014.

\bibitem[Hong and Nelson(2006)]{hong2006discrete}
L.~J. Hong and B.~L. Nelson.
\newblock Discrete optimization via simulation using compass.
\newblock \emph{Operations Research}, 54\penalty0 (1):\penalty0 115--129, 2006.

\bibitem[Huang et~al.(2006{\natexlab{a}})Huang, Allen, Notz, and
  Miller]{hanm06}
D.~Huang, T.~Allen, W.~Notz, and R.~Miller.
\newblock Sequential kriging optimization using multiple-fidelity evaluations.
\newblock \emph{Structural and Multidisciplinary Optimization}, 32\penalty0
  (5):\penalty0 369--382, 2006{\natexlab{a}}.

\bibitem[Huang et~al.(2006{\natexlab{b}})Huang, Allen, Notz, and
  Zeng]{HuAlNoZe06}
D.~Huang, T.~T. Allen, W.~I. Notz, and N.~Zeng.
\newblock {Global Optimization of Stochastic Black-Box Systems via Sequential
  Kriging Meta-Models}.
\newblock \emph{Journal of Global Optimization}, 34\penalty0 (3):\penalty0
  441--466, 2006{\natexlab{b}}.

\bibitem[Jones et~al.(1998)Jones, Schonlau, and Welch]{JoScWe98}
D.~R. Jones, M.~Schonlau, and W.~J. Welch.
\newblock {Efficient Global Optimization of Expensive Black-Box Functions}.
\newblock \emph{Journal of Global Optimization}, 13\penalty0 (4):\penalty0
  455--492, 1998.

\bibitem[Kandasamy et~al.(2016)Kandasamy, Dasarathy, Oliva, Schneider, and
  Poczos]{kdosp16}
K.~Kandasamy, G.~Dasarathy, J.~B. Oliva, J.~Schneider, and B.~Poczos.
\newblock Multi-fidelity gaussian process bandit optimisation.
\newblock In \emph{Advances in Neural Information Processing Systems}, 2016.

\bibitem[Kennedy and O'Hagan(2000)]{kh00}
M.~C. Kennedy and A.~O'Hagan.
\newblock Predicting the output from a complex computer code when fast
  approximations are available.
\newblock \emph{Biometrika}, 87\penalty0 (1):\penalty0 1--13, 2000.

\bibitem[Klein et~al.(2016)Klein, Falkner, Bartels, Hennig, and
  Hutter]{kfbhh16}
A.~Klein, S.~Falkner, S.~Bartels, P.~Hennig, and F.~Hutter.
\newblock Fast bayesian optimization of machine learning hyperparameters on
  large datasets.
\newblock \emph{CoRR}, abs/1605.07079, 2016.

\bibitem[Lam et~al.(2015)Lam, Allaire, and Willcox]{law15}
R.~Lam, D.~Allaire, and K.~Willcox.
\newblock Multifidelity optimization using statistical surrogate modeling for
  non-hierarchical information sources.
\newblock In \emph{56th AIAA/ASCE/AHS/ASC Structures, Structural Dynamics, and
  Materials Conference}, 2015.

\bibitem[LeCun et~al.()LeCun, Cortes, and Burges]{mnist}
Y.~LeCun, C.~Cortes, and C.~J. Burges.
\newblock The {MNIST} database of handwritten digits.
\newblock \url{http://yann.lecun.com/exdb/mnist/}. Last Accessed on 10/09/2016.

\bibitem[March and Willcox(2012)]{mw12}
A.~March and K.~Willcox.
\newblock Provably convergent multifidelity optimization algorithm not
  requiring high-fidelity derivatives.
\newblock \emph{AIAA Journal}, 50\penalty0 (5):\penalty0 1079--1089, 2012.

\bibitem[MOE()]{moe-github2015}
MOE.
\newblock Metrics optimization engine.
\newblock \url{http://yelp.github.io/MOE/}.
\newblock Last Accessed on 10/04/2016.

\bibitem[Picheny et~al.(2013)Picheny, Ginsbourger, Richet, and Caplin]{pgrc13}
V.~Picheny, D.~Ginsbourger, Y.~Richet, and G.~Caplin.
\newblock Quantile-based optimization of noisy computer experiments with
  tunable precision.
\newblock \emph{Technometrics}, 55\penalty0 (1):\penalty0 2--13, 2013.

\bibitem[Queipo et~al.(2005)Queipo, Haftka, Shyy, Goel, Vaidyanathan, and
  Tucker]{qhsgvt05}
N.~V. Queipo, R.~T. Haftka, W.~Shyy, T.~Goel, R.~Vaidyanathan, and P.~K.
  Tucker.
\newblock Surrogate-based analysis and optimization.
\newblock \emph{Progress in aerospace sciences}, 41\penalty0 (1):\penalty0
  1--28, 2005.

\bibitem[Rajnarayan et~al.(2008)Rajnarayan, Haas, and
  Kroo]{rajnarayan2008multifidelity}
D.~Rajnarayan, A.~Haas, and I.~Kroo.
\newblock A multifidelity gradient-free optimization method and application to
  aerodynamic design.
\newblock In \emph{Proceedings of the 12th AIAA/ISSMO Multidisciplinary
  Analysis and Optimization Conference}, number 6020, 2008.

\bibitem[Rasmussen and Williams(2006)]{rw06}
C.~E. Rasmussen and C.~K.~I. Williams.
\newblock \emph{Gaussian Processes for Machine Learning}.
\newblock MIT Press, 2006.
\newblock ISBN ISBN 0-262-18253-X.

\bibitem[Scott et~al.(2011)Scott, Frazier, and Powell]{sfp11}
W.~R. Scott, P.~I. Frazier, and W.~B. Powell.
\newblock The correlated knowledge gradient for simulation optimization of
  continuous parameters using gaussian process regression.
\newblock \emph{{SIAM} Journal on Optimization}, 21\penalty0 (3):\penalty0
  996--1026, 2011.

\bibitem[Snoek(2016)]{snoek16}
J.~Snoek.
\newblock Personal communication, 2016.

\bibitem[Snoek and et~al.()]{spearmint_github}
J.~Snoek and et~al.
\newblock Spearmint.
\newblock \url{http://github.com/HIPS/Spearmint}.
\newblock Last Accessed on 10/04/2016.

\bibitem[Swersky et~al.(2013)Swersky, Snoek, and Adams]{ssa13}
K.~Swersky, J.~Snoek, and R.~P. Adams.
\newblock Multi-task bayesian optimization.
\newblock In \emph{Advances in Neural Information Processing Systems}, pages
  2004--2012, 2013.

\bibitem[Teh et~al.(2005)Teh, Seeger, and Jordan]{tsj05}
Y.-W. Teh, M.~Seeger, and M.~Jordan.
\newblock Semiparametric latent factor models.
\newblock In \emph{Artificial Intelligence and Statistics 10}, 2005.

\bibitem[Theano()]{logregTheano}
Theano.
\newblock Theano: Logistic regression.
\newblock \url{http://deeplearning.net/tutorial/code/logistic_sgd.py}. Last
  Accessed on 10/08/16.

\bibitem[USPS()]{uspsdata}
USPS.
\newblock {USPS} dataset.
\newblock \url{http://mldata.org/repository/data/viewslug/usps/}. Last Accessed
  on 10/09/2016.

\bibitem[Villemonteix et~al.(2009)Villemonteix, Vazquez, and Walter]{vvw09}
J.~Villemonteix, E.~Vazquez, and E.~Walter.
\newblock An informational approach to the global optimization of
  expensive-to-evaluate functions.
\newblock \emph{Journal of Global Optimization}, 44\penalty0 (4):\penalty0
  509--534, 2009.

\bibitem[Winkler(1981)]{win81}
R.~L. Winkler.
\newblock Combining probability distributions from dependent information
  sources.
\newblock \emph{Management Science}, 27\penalty0 (4):\penalty0 479--488, 1981.

\bibitem[Xie et~al.(2012)Xie, Frazier, and Chick]{simopt_ato}
J.~Xie, P.~I. Frazier, and S.~Chick.
\newblock Assemble to order simulator.
\newblock
  \url{http://simopt.org/wiki/index.php?title=Assemble_to_Order&oldid=447},
  2012.
\newblock Last Accessed on 10/02/2016.

\end{thebibliography}
	\appendix
	
	\clearpage

\begin{center}
{ \Large 
	\misotitle	
}

{\large\textit{Supplementary Material}}
\end{center}

\section{The Model Revisited}
\label{section_model_appendix}

\subsection{Correlated Model Discrepancies}
\label{subsection_model_extension}
Next we demonstrate that our approach is flexible and can easily be extended to scenarios where some of the information sources have correlated model discrepancies.  This arises for hyper-parameter tuning if the auxiliary tasks are formed from data that was collected in batches and thus is correlated over time (see Sect.~\ref{section_intro} for a discussion). 
In engineering sciences we witness this if some sources share a common modeling approach, as for example, if one set of sources for an airfoil modeling problem correspond to different discretizations of a PDE that models wing flutter, while another set provides various discretizations of another PDE that modeling airflow. Two information sources that solve the same PDE will be more correlated than two that solve different PDEs.

For example, let~$P = \{P_1,\ldots,P_Q\}$ denote a partition of~$[M]_0$ and define the function~$k: [M]_0 \to [Q]$ that gives for each IS its corresponding partition in~$P$. 
Then we suppose an independent Gaussian process~$\varepsilon(k(\ell),x) \sim GP(\mu_{k(\ell)},\Sigma_{k(\ell)})$ for each partition.
(Note that in principle we could take this approach further to arbitrary sets of~$[M]_0$. However, this comes at the expense of a larger number of hyper-parameters that need to be estimated.)
Again our approach is to incorporate all Gaussian processes into a single one with prior distribution~$f \sim GP\left(\mu,\Sigma\right)$:\footnote{For simplicity we reuse the notation from the first model to denote their pendants in this model.} therefore, for all~$\ell \in [M]_0$ and~$x \in \domain$ we define~$f(\ell,x) = f(0,x) + \varepsilon(k(\ell),x) + \delta_\ell(x)$,
where~$f(0,x) = g(x)$ is the objective function that we want to optimize.
Due to linearity of expectation, we have~
\begin{align*}
\mu(\ell,x) & = \mathbb{E}\left[f(0,x) + \varepsilon(k(\ell),x) + \delta_\ell(x)\right]\\
& = \mathbb{E}\left[f(0,x)\right] + \mathbb{E}\left[\varepsilon(k(\ell),x)\right] + \mathbb{E}\left[\delta_\ell(x)\right]\\
& = \mu_0(x),
\end{align*}
since $\mathbb{E}\left[\varepsilon(k(\ell),x)\right] = \mathbb{E}\left[\delta_\ell(x)\right] = 0$.
Recall that the indicator variable~$\I_{\ell,m}$ denotes Kronecker's delta.
Let~$\ell,m \in [M]_0$ and~$x,x' \in \domain$, then we define the following composite covariance function~$\Sigma$: 
\begin{align*}
& \Sigma\left((\ell,x),(m,x')\right)\\
= \; & \mathrm{Cov}\left(f(0,x) + \varepsilon(k(\ell),x) + \delta_\ell(x), f(0,x')\right.\\
& \; \left. + \; \varepsilon(k(m),x') + \delta_m(x')\right)\\
= \; & \mathrm{Cov}( f(0,x), f(0,x')) + \mathrm{Cov}(\varepsilon(k(\ell),x),\varepsilon(k(m),x'))\\
& \; + \; \mathrm{Cov}(\delta_\ell(x), \delta_m(x'))\\
= \;  & \Sigma_0(x,x') + \I_{k(\ell),k(m)} {\cdot} \Sigma_{k(\ell)}(x,x') + \I_{\ell,m} {\cdot} \Sigma_\ell(x,x').
\end{align*}

\subsection{Estimation of Hyper-Parameters}
\label{section_fidelity_hypers}
In this section we detail how to set the hyper-parameters via \emph{maximum a posteriori}~(MAP) estimation and propose a specific prior that has proven its value in our application and thus is of interest in its own right. 

In typical MISO scenarios little data is available, that is why we suggest MAP estimates that in our experience are more robust than maximum likelihood estimates (MLE) under these circumstances.
However, we wish to point out that we observed essentially the same performances of the algorithms when conducting the Rosenbrock and Assemble-to-Order benchmarks with maximum likelihood estimates for the hyper-parameters.

In what follows we use the notation introduced in Sect.~\ref{section_model}.
One would suppose that the functions~$\mu_0(\cdot)$ and~$\Sigma_\ell(\cdot,\cdot)$ with~$\ell \in [M]_0$ belong to some parameterized class:
for example, one might set~$\mu_0(\cdot)$ and each~$\lambda_\ell(\cdot)$ to constants, and suppose that~$\Sigma_\ell$ each belong to the class of Mat\'{e}rn covariance kernels (cp.\ Sect.~\ref{section_experiments} for the choices used in the experimental evaluation).
The hyper-parameters are fit from data using \emph{maximum a posteriori}~(MAP) estimation;
note that this approach ensures that covariances between information sources and the objective function are inferred from data.

For a Mat\'{e}rn kernel we have to estimate~$d+1$ hyper-parameters for each information source (see next subsection): $d$ length scales and the signal variance.
We suppose a normal prior~${\cal N}\left(\mu_i, \sigma_i^2\right)$ for hyper-parameter~$\theta_i$.
Let~$D \in \domain$ be a set of points, for example chosen via a Latin Hypercube design, and evaluate every information source at all points in~$D$. 
We estimate the hyper-parameters for~$f(0,\cdot)$ and the~$\delta_i$ for~$i \in [M]$, using the ``observations''~$\Delta_i = \{\IS_i(x) - \IS_0(x) \; \mid \; x \in D\}$ for the~$\delta_i$.
The prior mean of the signal variance parameter of~$\IS_0$ is set to the variance of the observations at~$\IS_0$ minus their average observational noise. The mean for the signal variance of~$\IS_i$ with~$i \in [M]$ is obtained analogously using the ``observations'' in~$\Delta_i$; here we subtract the mean noise variance of the observations at~$\IS_i$ and the mean noise at~$\IS_0$, exploiting the assumption that observational noise is independent.
Regarding the means of the priors for length scales, we found it useful to set each prior mean to the length of the interval that the corresponding parameter is optimized over.
For all hyper-parameters~$\theta_i$ we set the variance of the prior to~$\sigma_i^2 = (\frac{\mu_i}{2})^2$, where~$\mu_i$ is the mean of the prior.

\subsection{How to Express Beliefs on Fidelities of Information Sources}
In many applications one has beliefs about the relative accuracies of information sources. 
One approach to explicitly encode these is to introduce a new coefficient~$\alpha_\ell$ for each~$\Sigma_\ell$ that typically would be fitted from data along with the other hyper-parameters.
But we may also set it at the discretion of a domain expert, which is particularly useful if none of the information sources is an unbiased estimator and we rely on regression to estimate the true objective.
In case of the squared exponential kernel this coefficient is sometimes part of the formulation and referred to as ``signal variance'' (e.g., see~\cite[p.~19]{rw06}).
For the sake of completeness, we detail the effect for our model of uncorrelated information sources stated in Sect.~\ref{section_simple_model}. Recall that we suppose~$f \sim GP(\mu,\Sigma)$ with a mean function $\mu$ and covariance kernel $\Sigma$, and observe that the introduction of the new coefficient~$\alpha_\ell$ does not affect~$\mu\left(\ell,x\right)$. But it changes~$\Sigma\left((\ell,x),(m,x')\right)$ to
\begin{equation*}
\Sigma\left((\ell,x),(m,x')\right) = \Sigma_0(x,x') + \I_{\ell,m} \cdot \alpha_\ell \cdot \Sigma_\ell(x,x').
\end{equation*}
We observe that setting~$\alpha_\ell$ to a larger value results in a bigger uncertainty. The gist is that then samples from such an information source have less influence in the Gaussian process regression (e.g., see Eq.~(A.6) on pp.~200 in~\cite{rw06}). 
It is instructive to consider the case that we observe a design~$x$ at a noiseless and deterministic information source: then its observed output coincides with~$f(\ell,x)$ (with zero variance). 
Our estimate~$f(0,x)$ for~$g(x)$, however, is a Gaussian random variable whose variance depends (in particular) on the uncertainty of the above information source as encoded in~$\alpha_\ell$, since~$\lambda_\ell(x) = 0$ holds.

	\section{Parallel Computation of the Cost-Sensitive Knowledge Gradient}
\label{section_parallel_KG_comp}
In Sect.~\ref{section_voi} we detailed how the cost-sensitive Knowledge Gradient can be computed. In particular, we discretized the inner optimization problem in Eq.~(\ref{Eq_KG_cost}) to obtain
\begin{multline*}
\E_n\left[\frac{\max_{x' \in \discretedomain}{\mu^{(n+1)}(0,x')} - \max_{x' \in \discretedomain}{\mu^{(n)}(0,x')}}{c_{\ell}(x)} \;\middle|\; \right. \\
\left. \ell^{(n+1)} = \ell,x^{(n+1)} = x\right].
\end{multline*}
Then we suggested exploiting the gradient of the CKG factor to obtain the next sample point and information source that maximize the expected gain per unit cost.

While we found this approach to work well in our experimental evaluation, there are scenarios where it is beneficial to also discretize the outer maximization problem~$\argmax_{\ell \in [M]_0, x \in \discretedomain}\mathrm{CKG}(\ell,x)$ and find the best~$x \in \discretedomain$ by enumerating all CKG factors, for example when the CKG over~$\domain$ has many local maxima and therefore gradient-based optimization techniques may fail to find the best sample location.
However, the running time of this approach is~$O\left(M \cdot |\discretedomain|^2 \cdot \log(|\discretedomain|)\right)$, and therefore may become a bottleneck in probing the domain with high accuracy.
We note in passing that the logarithmic factor can be shaved off by a suitable sorting algorithm that exploits the property that we are sorting numbers and hence runs in time~$O(|\discretedomain|)$.
In this section we propose a parallel algorithmic solution of essentially linear speed up, that makes efficient use of modern multi-core architectures.

We present two ideas to improve the scalability: the first stems from the observation that the computations for different choices of the next sample decision~$\left(\ell^{(n+1)},x^{(n+1)}\right)$ are independent and thus can be done in parallel; the only required communication is to scatter the data and afterwards determine the best sample, hence the speedup is essentially linear.
The second optimization is more intricate: we also parallelize the computation of the value of information for each~$\left(\ell^{(n+1)},x^{(n+1)}\right)$. Thus, we offer \emph{two levels of parallelization} that can be used separately or combined to efficiently utilize several multi-core CPUs of a cluster.

First recall that the query cost only depends on the choice of $\left(\ell^{(n+1)},x^{(n+1)}\right)$ and thus can be trivially incorporated; we omit it in the sequel for a more compact presentation.
Note that the same set of discrete points~$\discretedomain \subset \domain$ is used in the inner and the outer maximization problem only for the sake of simplicity; we could also use different sets and additionally exploit potential benefits of choosing them adaptively.
Moreover, let~$\rho : [M]_0 \times \discretedomain \to \left[(M+1) \cdot |\discretedomain|\right]$ be a bijection and define the~$((M+1) \cdot |\discretedomain|)$-dimensional vectors~$\bar{\mu}^{n}$ and~$\bar{\sigma}(\ell,x)$ as~$\bar{\mu}^{n}_{\rho(\ell',x')} = \mu^{(n)}(0,x')$ and~$\bar{\sigma}^n(\ell,x)_{\rho(\ell',x')} = \tilde{\sigma}^n_{x'}(\ell,x)$ respectively, where~$\tilde{\sigma}^n(\ell,x)$ is the analog of~$\tilde{\sigma}^n(x)$ in \citep{FrPoDa_Correlated} (see also Sect.~\ref{section_voi}).
Then we can define and using~$Z \sim {\cal N}(0,1)$
\begin{equation*}
h(\vec{a},\vec{b}) = \mathbb{E}\left[\max_i a_i + b_i \cdot Z\right] - \max_i a_i,
\end{equation*}
and thus observe that
\begin{align*}
& \E_n\left[\max_{x' \in \discretedomain}{\mu^{(n+1)}(0,x')} - \max_{x' \in \discretedomain}{\mu^{(n)}(0,x')} \;\middle|\; \right.\\
& \qquad \left. \ell^{(n+1)} = \ell, x^{(n+1)} = x\right]\\
= & h\left(\bar{\mu}^{n},\bar{\sigma}^n(\ell,x)\right).
\end{align*}
Comparing these ideas to~\citep{FrPoDa_Correlated},  the first modification corresponds to a parallel execution of the outer loop of Algorithm~2 of \citet{FrPoDa_Correlated} that computes the~$h$-function, whereas the second aims at parallelizing \emph{each} iteration of the loop itself.
Due to space constraints we only provide a sketch of the parallel algorithm here and assume familiarity with the algorithm of \citet{FrPoDa_Correlated}.

We begin its computation by sorting the entries of~$\bar{\mu}^{n}$ and~$\bar{\sigma}^n(\ell,x)$ \emph{in parallel} by ascending~$\bar{\sigma}^n(\ell,x)$-value; if multiple entries have the same~$\bar{\sigma}^n(\ell,x)$, we only keep the entry with largest value in~$\bar{\mu}^{n}$ and discard all others, since they are dominated~\citep{FrPoDa_Correlated}. W.l.o.g.\ we assume in the sequel that both vectors do not contain dominated entries and that~$\bar{\sigma}^n(\ell,x)_i < \bar{\sigma}^n(\ell,x)_j$ whenever~$i < j$ for~$i,j \in [(M+1) \cdot |\discretedomain|]$.
However, there is another type of domination that is even more important: for each~$z \in \mathbb{R}$ it is sufficient to find the~$g(z) := \max\argmax_i \bar{\mu}^{n}_i +  \bar{\sigma}^n(\ell,x)_i \cdot z$, which is equivalent to removing those~$i$ that never maximize~$\bar{\mu}^{n}_i +  \bar{\sigma}^n(\ell,x)_i \cdot z$ for any~$z$. Let~$n'$ be the number of sample points that remain after this step and consider the sequence~$(c)$ with~$c_i = \frac{\bar{\mu}^{n}_i - \bar{\mu}^{n}_{i+1}}{\bar{\sigma}^n(\ell,x)_{i+1} - \bar{\sigma}^n(\ell,x)_i}$ for~$i \in [n' - 1]$, $c_0 = -\infty$ and~$c_{n'} = \infty$. We observe that the intervals between these points uniquely determine the respective~$g(z)$ for all~$z$ in that interval.

In order to parallelize Algorithm~1 of~\citet{FrPoDa_Correlated} that determines un-dominated candidates for the next sample point (also called alternatives), we divide the previously sorted sequence~$(c)$ into~$p$ subsequences, where~$p$ is the number of cores we wish to use. 
After running the linear scan algorithm of~\citet{FrPoDa_Correlated} on each subsequence separately and in parallel, we ``merge'' adjacent subsequences pairwise in parallel: when merging two sequences, say~$L$ and~$R$, where~$L$ contains the smaller elements, it suffices to search for the rightmost element in~$L$ not dominated by the leftmost one in~$R$ (or vice versa). The reason is that we have ensured previously that no element is dominated by another within each subsequence.

After at most~$\lceil\log_2 p\rceil$ merging rounds, each core has determined which of the elements in its respective subsequence of~$(c)$ are not dominated as required by Algorithm~2. 
Note that the ``merging'' procedure does not require actually transfer the elements among cores.
Hence the final step, the summation in Eq.~(14) on p.~605 in~\citep{FrPoDa_Correlated} that calculates~$h\left(\bar{\mu}^{n},\bar{\sigma}^n(\ell,x)\right)$, is trivial to parallelize.

\end{document}